\documentclass[letterpaper, 10 pt, conference]{ieeeconf}  

\IEEEoverridecommandlockouts                              

\usepackage{graphics} 
\usepackage{graphicx}
\usepackage{times} 
\usepackage{amsmath} 
\usepackage{bm}
\usepackage[noadjust]{cite}
\usepackage{amssymb}  
\usepackage[linesnumbered,ruled,vlined]{algorithm2e}
\usepackage{svg}
\usepackage{soul}
\allowdisplaybreaks[4]
\svgpath{{fig/assets/}} 
\def\BibTeX{{\rm B\kern-.05em{\sc i\kern-.025em b}\kern-.08em
    T\kern-.1667em\lower.7ex\hbox{E}\kern-.125emX}}
\usepackage{todonotes}

\newcommand{\name}{DIPS }
\newcommand{\namenospace}{DIPS}

\title{\LARGE \bf
Diffusion-Informed Probabilistic Contact Search for Multi-Finger Manipulation
}

\author{Abhinav Kumar$^{*1}$, Thomas Power$^{*1}$, Fan Yang$^{1}$, Sergio Aguilera Marinovic$^{2}$,\\Soshi Iba$^{2}$, Rana Soltani Zarrin$^{2}$, Dmitry Berenson$^{1}$
\thanks{$^*$ Equal contribution. $^{1}$Robotics Department, 
        University of Michigan, Ann Arbor, MI, USA
        {\tt\small [abhin, tpower, fanyangr,dmitryb]@umich.edu},  $^{2}$ Honda Research Institute USA. This work was sponsored by Honda Research Institute USA.}%
}

\begin{document}

\maketitle
\thispagestyle{empty}
\pagestyle{empty}

\begin{abstract}
\looseness-1
Planning contact-rich interactions for multi-finger manipulation is challenging due to the high-dimensionality and hybrid nature of dynamics.
Recent advances in data-driven methods have shown promise, but are sensitive to the quality of training data. 
Combining learning with classical methods like trajectory optimization and search adds additional structure to the problem and domain knowledge in the form of constraints, which can lead to outperforming the data on which models are trained.
We present Diffusion-Informed Probabilistic Contact Search (DIPS), which uses an A* search to plan a sequence of contact modes informed by a diffusion model.
We train the diffusion model on a dataset of demonstrations consisting of contact modes and trajectories generated by a trajectory optimizer given those modes.
In addition, we use a particle filter-inspired method to reason about variability in diffusion sampling arising from model error, estimating likelihoods of trajectories using a learned discriminator.
We show that our method outperforms ablations that do not reason about variability and can plan contact sequences that outperform those found in training data across multiple tasks. 
We evaluate on simulated tabletop card sliding and screwdriver turning tasks,
as well as the screwdriver task in hardware to show that our combined learning and planning approach transfers to the real world.

\end{abstract}

\vspace{-.2cm}
\section{Introduction}
\vspace{-.1cm}
Multi-finger manipulation is challenging as there are many ways in which the hand can make or break contact with manipulated objects.
The discrete modes in which the hand makes contact induce different dynamics by imposing different constraints on the system.
Thus the system is hybrid and there is no clear way to sequence contact modes in order to accomplish a given task efficiently.

Recent advances in learning methods, specifically generative modeling \cite{trajdiffuser}, \cite{chi_diffusion_2024}, can be used to learn manipulation policies without requiring strong domain knowledge of the task in the form of constraint modeling.
However, in multi-finger manipulation, which has sensitive constraints relating to contact, that domain knowledge is valuable to improve task performance.
In addition, learning methods generate policies similar to the demonstration data on which they are trained.
In tasks with varied possible contact interactions, a planned contact mode sequence could yield better results than that used in data collection.

In contrast to learning-based methods, planning methods that use techniques such as trajectory optimization offer a way to reason about constraints.
While these methods can be used to plan a trajectory given a specific contact mode \cite{fanrolling} or can plan contact mode changes for lower-dimensional systems \cite{natarajan_torque-limited_2023, posa2014direct}, planning contact sequences in high-dimensional multi-finger manipulation problems can become intractably expensive to compute.

We approach multi-finger manipulation by combining the flexibility of learning with the constraint satisfaction of model-based methods.
In our method, Diffusion-Informed Probabilistic Contact Search (DIPS), we train a diffusion model \cite{ddpm} on trajectories generated by a trajectory optimizer.
We then use this model to generate trajectories corresponding to edges in an A* search over contact modes.

We also propose a method to reason about the variability in the diffusion sampling, which can cause issues in contact sequence planning.
This variability could come from low quality samples arising from the innate error of learned models or the quality of the training data.
Reasoning about variability in the diffusion output can be helpful when planning sequences of multiple contact modes in which multiple diffusion samples would need to be concatenated over time.

To reason about the variability in diffusion sampling when planning contact sequences, we use a particle-based approximation of the distribution over trajectories modeled by the diffusion.
We train an additional discriminator that assigns higher scores to more realistic trajectories to provide likelihood estimates of particles.
Using the diffusion sampling and likelihood estimation, we construct a particle-based method to propagate variability estimates as edges are sequenced in the search. Our contributions are:

\begin{figure}[t]    
    \centerline{\includesvg[inkscapelatex=false, scale=.7]{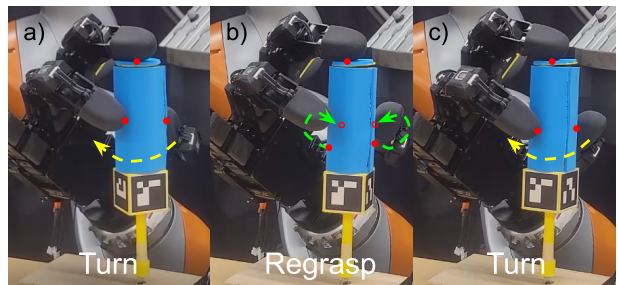}}    
    \vspace{-0.3cm}
    \caption{\name plans contact interactions that turn the screwdriver from \textbf{a)} to \textbf{b)}, where it is regrasped  to allow for further turning in \textbf{c)}. Contact points are shown in red, with empty circles for target contacts. The yellow arrows show screwdriver turning and green show finger motion. 
    }
    \label{fig:title}
\end{figure}

\begin{itemize}
    \item A method for planning contact mode sequences using graph search informed by a diffusion model
    \item A method for explicitly reasoning about the variability in the diffusion model sampling when planning
\end{itemize}

We show that our method can be applied to challenging tool-use tasks and that \name outperforms ablations and baselines that do not plan contact sequences and that do not reason about the variability in the diffusion model output.

\vspace{-.1cm}
\section{Related Work}
\vspace{-.1cm}

\paragraph{Planning for Contact-Rich Manipulation}
Due to the difficulty of planning through contact, recent work has optimized with a pre-defined contact mode \cite{fanrolling}, contact schedule \cite{sleiman_whole_body_mpc}, or sequence of constraints \cite{toussaint2022sequenceofconstraintsmpcreactivetimingoptimal}. Other work has explicitly planned contact sequences in sample-based motion planning frameworks \cite{fingergaiting, cheng2022contactmodeguidedmotion}. Smooth contact models have also been used to simplify the planning problem. These have been used for trajectory optimization \cite{rozzi2024combining, pang_smoothed_contact} as well as sample-based planners \cite{pang_smoothed_contact}. 
\paragraph{Combining Search \& Trajectory Optimization}
Several recent works have combined search and trajectory optimization. Cheng et al. propose a hierarchical approach based on Monte-Carlo Tree Search to explore contact modes \cite{cheng_contact_exploration}. Other recent work has used tree-based planners to explore contact modes combined with contact-implicit trajectory optimization \cite{Chen2021TrajectoTreeTO, zhang2023simultaneous}.
There have been several recent works combining A* search with trajectory optimization \cite{natarajan_interleaving_2021, natarajan_torque-limited_2023, natarajan_pinsat_2024, chia_gcs_2024}. Unlike previous work combining A* and trajectory optimization, we accelerate our planning using a learned model to approximate trajectory optimization.  
\paragraph{Learning for Contact-Rich Manipulation}
In recent years learning-based methods have been increasingly popular for solving contact-rich tasks. These include methods that learn models for planning \cite{kumar_learning_2016, PDDM}, as well as methods that learn policies using reinforcement learning \cite{popov_data-efcient_nodate, zhu_dexterous_2019, huang_generalization_2021, xu_dexterous_2023}. In general, these methods can solve complex tasks when given sufficient training data but often require large amounts of training data. Imitation learning has also been applied to dexterous manipulation, accelerating learning with demonstrations \cite{rajeswaran_learning_2018, shaw_dexterity_from_videos}. These methods are effective but rely on collecting high-quality demonstrations. 
Efforts to combine learning and classical approaches have used reinforcement learning to learn mid-level policies to sequence primitives or controllers \cite{li_hierarchical, khandate_subcontrollers, zarrin2023hybrid, gordan_hybrid}.
Our method similarly aims to combine learning with planning, but can automatically generate high-quality demonstrations from trajectory optimization.

\begin{figure*}[t]
\label{fig:block}
\centerline{\includesvg[inkscapelatex=false, scale=.52]{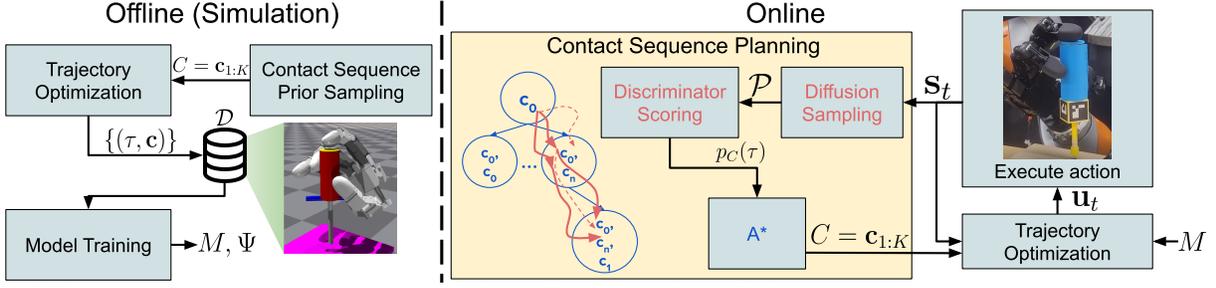}}
\vspace{-.35cm}
\caption{
\looseness-1
\textbf{Offline}, we sample contact sequences from a designed prior.
We generate a dataset $\mathcal{D}$ of trajectories in simulation.
We train a diffusion model $M$ and discriminator $\Psi$ on $\mathcal{D}$.
\textbf{Online}, we plan a contact sequence $C$ given a state $\mathbf{s}_t$.
We expand nodes in blue corresponding to contact mode sequences and inform the search using a distribution $p_C(\bm{\tau})$ parameterized with a set of trajectories $\mathcal{P}$ in pink.
We diffuse trajectories conditioned on the child node's contact mode and evaluate them with $\Psi$.
The dotted lines are samples discarded in the resampling used to update $p_C(\bm{\tau})$.
Given a single contact mode and $\mathbf{s}_t$, we optimize a trajectory of length $H$ initialized with samples from $M$.
We rerun the trajectory optimization every timestep.
After each contact mode, we replan $C$.
}
\vspace{-.7cm}
\end{figure*}
\vspace{-.2cm}
\section{Problem Statement}
\vspace{-.1cm}

In this paper, we consider the problem of contact-rich manipulation with a multi-fingered hand. Our goal is to find a sequence of robot configurations $\mathbf{q}_{1:T}$, where $T$ is the number of execution timesteps, that successfully manipulate an object from start pose $\mathbf{o}_0$ to goal pose $\mathbf{o}_G$.
This task is challenging as successfully manipulating the object may require the making and breaking of contact, resulting in non-smooth dynamics. 
We define contact mode $\mathbf{c}:=\{0, 1\}^{n_f}$, where $n_f$ is the number of fingers. $\mathbf{c}$ is a binary vector specifying which finger should be in contact. Given a contact mode, we formulate the following trajectory optimization problem for horizon $H < T$.

\vspace{-.4cm}
\begin{align}
\begin{split}
    \mathbf{s}^*_{1:H}, \mathbf{u}^*_{1:H} &= \arg \min J(\mathbf{s}_{1:H}, \mathbf{u}_{1:H}, \mathbf{c}) \\
    &\text{s.t.} \quad   
    h(\mathbf{s}_{1:H}, \mathbf{u}_{1:H}, \mathbf{c}) = 0 \\
    &g(\mathbf{s}_{1:H}, \mathbf{u}_{1:H}, \mathbf{c}) \leq 0,
\end{split}
\label{ps:trajopt_segment}
\end{align}
with state $\mathbf{s}_t$, action $\mathbf{u}_t$ and cost and constraint functions $J, h, g$, which are all dependent on the contact mode. For shorthand, we denote the trajectory $\bm{\tau}:= \{\mathbf{s}_{1:H}, \mathbf{u}_{1:H}\}$. Once the contact mode is provided, this trajectory optimization problem can be efficiently solved. To make this problem tractable, we assume (1) access to the geometries of the hand and the object, (2) that the system is quasi-static, and (3) that the fingers begin the task in contact with the object.

To allow for the making and breaking of contact, we split the trajectory into $K$ segments, i.e. $H = \frac{T}{K}$, and $\bm{\tau} = \{\bm{\tau}_1, ..., \bm{\tau}_K\}$. The aim is then to find a sequence of contact modes $C=\mathbf{c}_{1:K}$, with $\bm{\tau_k}$ being the solution to Problem (\ref{ps:trajopt_segment}) with contact mode $\mathbf{c}_k$ for $k \in [1, ..., K]$, such that the overall trajectory $\bm{\tau}$ results in completing the task.

To aid in solving this problem we assume we can generate a dataset of $N$ trajectories, $\mathcal{D} = \{\bm{\tau}_i, \mathbf{c}_i\}^N_{i=1}$ which consists of solutions to Problem ($\ref{ps:trajopt_segment}$) for diverse initial configurations.

We evaluate our method on its ability to achieve the goal $\mathbf{o}_G$.
For some tasks, we also consider the validity of the executed trajectories.
For example, if the hand is manipulating a screwdriver, dropping the screwdriver would be the consequence of an invalid trajectory.

\vspace{-.2cm}
\section{Methods}
In this section, we outline our method for in-hand contact-rich manipulation that combines search and trajectory optimization to search for a sequence of contact modes $\{\mathbf{c}_1, ..., \mathbf{c}_K\}$ and optimize trajectories given these modes.

Our method, shown in Fig. 2, uses A* to search for the contact sequence. During planning, when expanding a potential contact mode $\mathbf{c}_k$, we must reason about the resulting trajectory $\bm{\tau}_k$. Since solving the trajectory optimization problem in the inner loop of our planner would be prohibitively expensive, we instead train a diffusion model on a dataset of trajectories and sample from this as a proxy for solving the full optimization.
Finally, we optimize and execute trajectories given the planned contact sequences using the trajectory optimizer.
We also use the samples from the diffusion model to initialize the trajectory optimization.

\vspace{-.1cm}
\subsection{Trajectory Optimization} \label{sec:traj_opt}
Our trajectory optimization formulation is based on prior work by Yang et al. \cite{fanrolling}. The formulation in \cite{fanrolling} assumes a fixed contact mode and only optimizes the motion for fingers in contact. We extend this formulation to be conditioned on the contact mode and additionally optimize the motion of specified fingers so they can ``regrasp'' i.e., make contact in a different location. 

The state $\mathbf{s}$ 
consists of finger configurations $\{\mathbf{q}_i\}_{i=1}^{n_f}$ and object pose $\mathbf{o}$.
The control vector $\mathbf{u}$ is $\{\{\Delta \mathbf{q}_i, \mathbf{f}_i\}^{n_f}_{i=1}, \mathbf{f}_e\}$, where $\mathbf{f}_i$ is the contact force for the $i$th finger, and $\mathbf{f}_e$, the contact force exerted by the environment. All contact forces are defined in the object frame. 
Given a contact mode $\mathbf{c} \in \{0, 1\}^{n_f}$ we partition the state and control vectors into contact fingers $\{\mathbf{s}_c, \mathbf{u}_c\} = \{\mathbf{q}_i, \Delta \mathbf{q}_i, \mathbf{f}_i : \mathbf{c}_i = 1\}$, regrasping fingers $\{\mathbf{s}_r, \mathbf{u}_r\} = \{\mathbf{q}_i, \Delta \mathbf{q}_i : \mathbf{c}_i = 0\}$, and the object and environment $\{\mathbf{s}_o, \mathbf{u}_o\} = \{\mathbf{o}, \mathbf{f}_e\}$. 
There is no contact force for regrasping fingers, as they break contact.
We similarly partition the trajectory into $\bm{\tau} = \{\bm{\tau}_c, \bm{\tau}_r, \bm{\tau}_o\}$. Our trajectory optimization problem is then written as
\vspace{-.45cm}

\begin{equation}
\begin{aligned}
& \min_{\substack{\mathbf{s}_1, \mathbf{s}_2, \cdots, \mathbf{s}_H; \\ \mathbf{u}_1, \mathbf{u}_1, \cdots, \mathbf{u}}_H}  J_{g}(\bm{\tau_o}) + + J_{r}(\bm{\tau_r}, \bm{\tau_o}) + J_{smooth}(\bm{\tau}) \\
&\text{s.t.} \quad \mathbf{q}_{min} \leq \mathbf{q}_t \leq \mathbf{q}_{max} \\
&\mathbf{u}_{min} \leq \mathbf{u}_t \leq \mathbf{u}_{max}\\
&f_{contact}(\mathbf{s}_{c,t}, \mathbf{s}_{o, t}) = 0 \\
&f_{kinematics}(\mathbf{s}_{c,t}, \mathbf{s}_{o,t}, \mathbf{s}_{c, t+1}, \mathbf{s}_{o, t+1}) = 0 \\
&f_{balance}(\mathbf{s}_{c, t}, \mathbf{s}_{o, t},\mathbf{s}_{c, t+1}, \mathbf{s}_{o, t+1} \mathbf{u}_{c,t}, \mathbf{u}_{o,t}) = 0\\
& f_{friction}(\mathbf{s}_{c,t}, \mathbf{s}_{o,t}, \mathbf{u}_{c,t}) \leq 0 \\
&f_{contact}(\mathbf{s}_{r,t}, \mathbf{s}_{o,t}) \leq  - \delta , t < H \\
&f_{contact}(\mathbf{s}_{r,H}, \mathbf{s}_{o,H}) = 0 \\
&\mathbf{q}_{r, t} + \Delta \mathbf{q}_{r,t} - \mathbf{q}_{r, t+1}  = 0.\\
\end{aligned}
\end{equation}
\vspace{-.3cm}

The cost term $J_g$ encourages the object to reach the goal location, $J_{smooth}$ incentivizes a smooth trajectory, and $J_r$ is a cost on the distance to target contact points for the regrasping fingers. The constraints $f_{kinematics}$, $f_{contact}$, $f_{balance}, f_{friction}$ are unchanged from \cite{fanrolling}. 
$f_{contact} \leq - \delta$ ensures that the regrasping fingers avoid contact with a threshold $\delta$ up until the final time step. The final constraint ensures that configurations and actions are consistent for the regrasping fingers that move in freespace.

The target contact points for the regrasping can be defined based on the task.
For example, if turning a screwdriver, we can set the targets to be the initial contact points of the fingers on the screwdriver to be able to reset fingers after turning.
Other tasks may benefit from other specifications.

To solve the trajectory optimization we use Constrained Stein Variational Trajectory Optimization (CSVTO) \cite{power2024constrained}. 
This optimization formulation allows us to generate trajectories given a pre-specified contact mode. We will next discuss how we use this to generate high-quality demonstrations used to train a diffusion model for a variety of contact modes. 

\vspace{-.2cm}
\subsection{Diffusion Model Training}
\label{sec:diffusion_training}

To aid in contact sequence planning, we train a diffusion model $M(\mathbf{c}, \mathbf{s}_0)$.
We can use $M$ to sample from the distribution $p(\bm{\tau}|\mathbf{c}, \mathbf{s}_0)$.
$p(\bm{\tau}|\mathbf{c}, \mathbf{s}_0)$ is the distribution modeling trajectories computed by the trajectory optimizer given a contact mode and initial state.
We train this model on a dataset $\mathcal{D}$ of trajectories and corresponding contact modes.

When generating data, we use a high optimization budget for CSVTO to optimize high-quality trajectories.
We can then optionally use diffusion samples when executing the task to initialize CSVTO, potentially requiring a lower optimization budget at runtime due to the higher-quality initialization.

To obtain the contact sequences used to generate $\mathcal{D}$, we sample from a constructed prior $p(C)$.
$p(C)$ is designed to accomplish a specific task, for example, turning a screwdriver.
While $p(C)$ is useful for generating data and should represent a reasonable attempt to solve the task, we find that our method can plan contact sequences that outperform $p(C)$.

We adopt the 1-D U-Net architecture used in \cite{trajdiffuser, power2023sampling} for the diffusion model.
We diffuse a trajectory of dimension $H \times (d_s + d_u)$, where $d_s$ is the dimensionality of the state, and $d_u$ is the dimensionality of the action.
We use classifier-free guidance \cite{ho2022classifierfreediffusionguidance} to condition on a specific contact mode $\mathbf{c}$.
To condition on $\mathbf{s}_0$, we use the same inpainting approach as \cite{trajdiffuser}.
At training time, we randomly sample masks over the trajectory, emulating inpainting masks, which we find improves the inpainting performance when sampling.

As we are working with complex systems with high degrees of freedom and complex constraints, $M$ may diffuse unrealistic trajectories.
As we will discuss in Section \ref{sec:seq_plan}, we compute weights of trajectories in our variability propagation method that represent their realism to account for this.
One way to compute these weights could be using the likelihood of the trajectory under the diffusion model, as shown in \cite{zhou2023adaptiveonlinereplanningdiffusion}.
However, we find using the likelihood to be intractable due to its high cost of approximation.

Instead, we train a discriminator $\Psi(\bm{\tau}, \mathbf{c})$ that takes in a trajectory and contact mode and outputs the probability that $\bm{\tau}$ is ``real'', or similar to $\mathcal{D}$. 
To train the discriminator, we use a dataset consisting of $\mathcal{D}$ and an equal number of trajectories sampled from the diffusion model.
We use a U-Net with a Sigmoid activation output layer to model $\Psi$.

\begin{algorithm}[t]
Given $\mathbf{n}^p=(\mathcal{P}^p$, $S^p, C^p)$, $\mathbf{c}'$, $M$, $\Psi$, $k$, $\gamma$\\
$\bar{\bm{\tau}} \gets$ Diffuse $k$ trajectories from $M(\mathbf{c}', \mathcal{P}^p_H)$\\
$d \gets \texttt{depth}(\mathbf{n}^p)$\\
$S \gets \gamma^d \cdot \Psi(\bar{\bm{\tau}}, \mathbf{c}') + S^p$\\
$\bar{S} \gets \{S_i / \sum\limits_{S_i \in S} S_i~|~\forall S_i \in S\}$\\
$\mathcal{P} \gets$ $k$ samples from $\bar{\bm{\tau}}$ given probabilities $\bar{S}$\\
$C \gets C^p \cup \mathbf{c}'$\\
$\mathbf{n} = (\mathcal{P}, S, C)$ \\
return $\mathbf{n}$
\caption{\texttt{propagate\_variability}}
\label{alg:uncertain_prop}
\end{algorithm}

\vspace{-.2cm}
\subsection{Probabilistic Contact Sequence Search}
\label{sec:seq_plan}
While using $C \sim p(C)$ can lead to task success, there are multiple reasons why planning contact modes can be beneficial.
First, there may be redundancies in the contact sequences used offline.
For example, finger positions may be reset more often than is necessary.
In addition, planning contacts reduces the reliance on a strong prior for achieving good task performance.

To find the contact mode sequence, we construct an A* tree search problem where each node $\mathbf{n}$ in the tree corresponds to a contact mode sequence $C$.
The descendants of a parent node $\mathbf{n}^p$ are computed by appending an additional contact mode $\mathbf{c}'$ to the parent node's sequence $C^p$.

We use trajectories conditioned on $C$ to compute costs for A* and check if $\mathbf{o}_G$ has been achieved in the search.
As opposed to prior work \cite{natarajan_torque-limited_2023} that uses a single trajectory at each node, we model a distribution $p_C(\bm{\tau})$ at each node to reason about variability in diffusion sampling.
We parameterize $p_C(\bm{\tau})$ with a set of particles $\mathcal{P}$ where each particle is a trajectory diffused by $M$.
Each particle has a weight, calculated by normalizing a score $S$ computed by the discriminator $\Psi(\bm{\tau}, \bm{c})$.
The full definition of a node is $\mathbf{n} = (\mathcal{P}, S, C)$.
By using this particle-based representation, we can approximate $p(\bm{\tau}|\mathbf{c}, \mathbf{s}_0)$ with $p_C(\bm{\tau})$.

To expand to new nodes in the A* search and compute costs, we use our diffusion model.
As diffusion models are learned and therefore can be unreliable, it is possible to diffuse trajectories that are unrealistic.
To address this, we explicitly reason about the variability in the diffusion model output during the planning process.

\looseness-1
As shown in Algorithm \ref{alg:uncertain_prop}, we diffuse $k$ trajectories to construct a population $\bar{\bm{\tau}}$ from which we sample new particles.
We sample 1 trajectory for each particle from the parent node, conditioned on the child node's new contact mode $\mathbf{c}'$ and the endpoints of the parent trajectories $\mathcal{P}^p_H$ to enforce continuity.
We compute the weights $S$ for $\bar{\bm{\tau}}$ using $\Psi$.
We accumulate the scores as we expand the tree, discounted by a factor $\gamma^d$, where $d$ is the depth of the node in the search tree.
We sample $k$ trajectories from $\bar{\bm{\tau}}$ using the normalized scores.

Costs and goal evaluations for the search are calculated using expectations over the $k$ particles.
Combined with replanning after each contact mode is executed, explicitly reasoning about variability allows us to reduce the stochasticity of planned contact sequences.
We seek to compute a contact sequence that leads to a minimum cost trajectory to the goal.
We therefore design our cost-to-come $g$ using the CSVTO cost $J$ as shown in \eqref{eq:g}.

\vspace{-.3cm}
\begin{equation} \label{eq:g}
    g(p_C(\bm{\tau}), C) = \mathbb{E}_{\bm{\tau}\sim p_C(\bm{\tau})}[J(\bm{\tau}, C)]
\vspace{-.2cm}
\end{equation}
We take an expectation over particles weighted by their scores and pass in $C$ to account for mode-specific objectives.

A* uses an additional heuristic $h$, to guide the search and improve search speed.
We design a heuristic that focuses the search on contact sequences with high likelihood under the prior and is biased toward trajectories that have a lower goal cost.
To compute $h$, we compute an approximation of the likelihood of a contact sequence under $p(C)$ and also use a terminal cost $\phi(\bm{\tau})$ as used in model predictive control methods.
For example, by considering distance of the terminal state to the goal, we can encourage contact sequences that more quickly reach the goal.

We use a 1-step Markov approximation $p(\mathbf{c}_n|\mathbf{c}_{n-1})$ of $p(C)$ as a prior to guide the search.
However, there will be contact mode transitions that may not be present in $\mathcal{D}$ that we wish to consider when planning.
To address this, we enforce a minimum probability $p_{min}$ for transitions.

In our heuristic, shown in \eqref{eq:heuristic}, we add the first term, the expected terminal cost, to the second term, the negative log-likelihood of $C$ under the prior.
The terms are weighted by $\alpha, \beta \in \mathbb{R}$ respectively.
While not admissible, this heuristic is useful to guide our search and improve its efficiency.
\vspace{-.2cm}
\begin{multline}\label{eq:heuristic}
h(p_C(\bm{\tau}), \mathbf{c}_{1:N}) = \alpha \cdot \mathbb{E}_{\bm{\tau}\sim p_C(\bm{\tau})}[\phi(\bm{\tau})]\displaystyle\\
-\beta \cdot \left[ \mathrm{log}(p(\mathbf{c}_0)) + \sum_{n=1}^{N} \mathrm{log}(p(\mathbf{c}_n|\mathbf{c}_{n-1})) \right]
\end{multline}

\section{Experiments and Results}
We evaluate \name on 3 tasks using an Allegro multi-fingered hand: A simulated task in which the hand slides a card-like object along a table, a simulated task in which the hand turns a screwdriver, and the screwdriver-turning task in the real world.
In all tasks, the pose of the base of the hand is fixed.
Simulations are implemented in Isaac Gym \cite{makoviychuk2021isaacgymhighperformance}.

For each task, we define contact modes where each mode specifies objectives and constraints for CSVTO.
$\mathbf{o}_G$ specifies the goal of the A* planner, but the goal for a specific contact mode used in CSVTO will differ as we are attempting to achieve $\mathbf{o}_G$ through a series of contact interactions.
We define separate goals used with CSVTO for each contact mode.
In addition, we use a timeout of 300 seconds when running the A* search.
If the search times out, we return the node that most closely reaches $\mathbf{o}_G$.
For all tasks, we use $\delta = 0.015 \:m, k=16, \gamma=.9, \alpha=1 \times 10^4, \beta=1 \times 10^3$.

\begin{figure}[t]
    \centering\centerline{\includesvg[inkscapelatex=false, scale=.5]{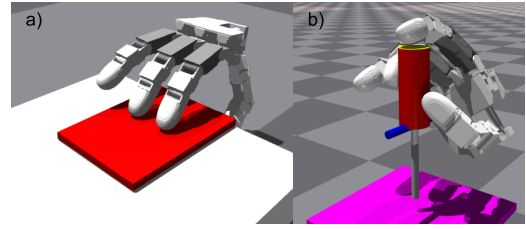}}
    \vspace{-0.3cm}
    \caption{\textbf{a)} Simulated card and \textbf{b)} Simulated screwdriver environments.
    The blue valve in b) is for visualization only and has no collision geometry.
    }
    \label{fig:sim_envs}
\end{figure}

\vspace{-.1cm}
\subsection{Ablations and Baselines}
We evaluate multiple ablations and baselines, running 10 trials for each method.
Optimization budgets are the same for \name and all ablations.
We run 5 ablations:
(1) ``CSVTO-Sampled Fixed Sequence'': We use $C\sim p(C)$ and initialize the trajectory optimizer as in \cite{fanrolling}; (2) ``\namenospace-Sampled Fixed Sequence'': We use $C\sim p(C)$ and samples from $M$ to initialize CSVTO; (3) ``\namenospace-No Contact Replanning'': We plan $C$ once at the beginning of the task, executing without replanning.
(4) ``\namenospace-No variability Propagation'': We use \name with $k=1$, thus removing variability propagation in the A* search; (5) ``\namenospace-Max likelihood'': We store a single trajectory at each node but diffuse multiple samples, picking the highest-scoring when expanding.

We also baseline our method against Diffusion Policy \cite{chi_diffusion_2024}, which uses a diffusion model to learn a receding horizon policy from demonstrations.
We use the simulated demonstrations to train diffusion policy, concatenating trajectories from different contact modes.
With this baseline, we seek to show the benefits of reasoning independently about different contact modes as well as the trajectory optimization.

While sample-based methods like Model Predictive Path Integral Control (MPPI) \cite{mppi}, might be considered, we do not include them here.
MPPI does not strictly enforce constraints involving contact, and prior work \cite{fanrolling} has shown it to struggle to perform dexterous tasks like screwdriver turning. 

\vspace{-.2cm}

\subsection{Simulated Tabletop Card Manipulation}
In this task, shown in Fig.~\ref{fig:sim_envs}a, the hand manipulates a card on a table.
We use $\mathbf{o}=[x, y, \theta]$, where $x,y$ are positions of the card in the world frame and $\theta$ is the card's yaw angle.
The goal is to use the index and middle fingers to slide the card -6 cm along the world $y$-axis toward the palm to set up a grasp.
We only plan the sliding behavior, not the grasping.
We report distance to the goal, also used for $\phi(\bm{\tau})$.

We define 4 contact modes: (a) the index finger moves the card while the middle finger regrasps, (b) the middle finger moves the card while the index finger regrasps, (c) both fingers move the card along the table, and (d) where both fingers regrasp.
The modes differ in the goal for CSVTO.
For (a), (b), and (c), the CSVTO goal is to move the card -2 cm along the world y axis toward the palm.
For (d), the CSVTO goal is to keep the card stationary.
We use a uniform prior for $p(C)$ in which all mode transitions are equally likely and generate 480 demonstrations, each with 5 contact modes.

We execute a maximum of 5 contact modes.
We run the A* planner before each contact mode, with a maximum depth that decreases as we execute contact modes to improve convergence to the goal.

\begin{figure}[t]
    \centering
    \centerline{\includesvg[inkscapelatex=false, scale=1.0]{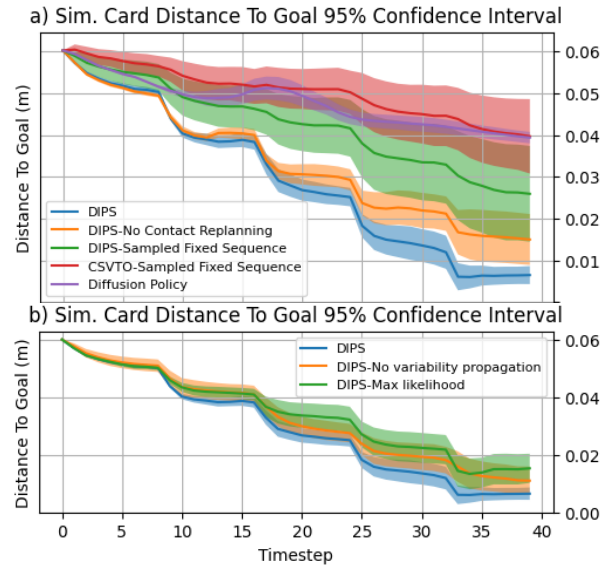}}
        \vspace{-0.3cm}
    \caption{Simulated Card  results over 10 trials.
    }
    \label{fig:card_results_plots}
\end{figure}

As shown in Fig.~\ref{fig:card_results_plots}(a,b), \name outperforms the baselines and ablations by avoiding unneeded regrasps. 
We come within .6 cm of $\mathbf{o}_G$ while \namenospace-Sampled Fixed Sequence comes within 2.6 cm and CSVTO-Sampled Fixed Sequence and Diffusion Policy come within 4 cm of $\mathbf{o}_G$.
Even though the training data sequences are uniformly randomly sampled, \name is still able to consistently produce useful contact sequences through the use of planning.
This shows the benefit of planning as a way to outperform training demonstrations.

\looseness-1
Diffusing initializations for CSVTO takes 3.6 s on average and CSVTO takes 7 s per step, while Diffusion policy takes 1.1 s.
Each A* planning call takes 64.9 s on average for \namenospace, 301.9 s for ablation (3), 52.7 s for ablation (4), and 13.8 s for ablation (5).
Expanding an edge with CSVTO would take approximately 18x long as using $M$, motivating the use of $M$ in the search.

\vspace{-.2cm}
\subsection{Simulated Screwdriver Turning}

In this task, shown in Fig.~\ref{fig:sim_envs}b, the hand turns a screwdriver using the thumb, index, and middle fingers.
The base of the screwdriver is attached to the table but can rotate, simulating driving a screw in a slot.
We define $\mathbf{o}$ as the orientation of the screwdriver, parameterized by its roll, pitch, and yaw.
The goal is to turn the screwdriver as far clockwise as possible.
For the A* goal $\mathbf{o}_G$ and $\phi(\bm{\tau})$, we only consider the yaw angle.
Additionally, because of the overall goal of turning the screwdriver as far as possible, we update $\mathbf{o}_G$ before each planning call.
Before planning, we set $\mathbf{o}_G$ to be $\frac{\pi}{3}$ less than the current yaw.
This is based on expecting to turn approximately $\frac{\pi}{2}$ across 7 modes in the prior, but a search over different values led to the optimal setting of $\frac{\pi}{3}$.
For \namenospace-No Contact Replanning, we performed a grid search to arrive at a goal of -1.7 rad.

We define 3 contact modes: (a) all 3 fingers are in contact and the hand is turning the screwdriver, (b) the thumb and middle finger are in contact and the index finger regrasps, and (c) the index finger is in contact and the thumb and middle fingers regrasp.
For (a), the CSVTO goal is to maintain the same roll and pitch while reducing the yaw by $\frac{\pi}{6}$.
For (b) and (c), the goal is to maintain the same screwdriver pose.

To sample from $p(C)$, we sequence (a), followed by (b) and (c) in random order, then repeat.
This means we turn, then regrasp all fingers, randomly ordering the regrasp modes, then resume turning.
We calculate a Markov approximation of the prior with $p_{min}=.1$:  $p(\mathbf{c}_0) = \begin{bmatrix}
        .1 & .1 & .8 \\
    \end{bmatrix}, p(\mathbf{c}_n|\mathbf{c}_{n-1}) = .1$ if $c_n=c_{n-1}$, .45 otherwise.
We generate 240 training demonstrations, each with 7 modes.

Online, we execute a maximum of 7 modes.
We use the same depth of 7 for A* throughout the task to encourage turning as far as possible.

\begin{figure}[t]
    \centering
    \centerline{\includesvg[inkscapelatex=false, scale=1.0]{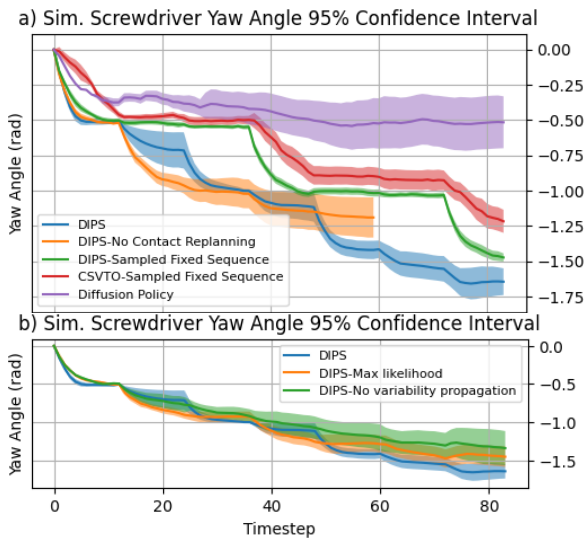}}
        \vspace{-0.3cm}
    \caption{Simulated Screwdriver results over 10 trials.
    }
    \label{fig:screwdriver_results_plots}
\end{figure}

As shown in Fig.~\ref{fig:screwdriver_results_plots}(a,b), \name outperforms the ablations and baselines, turning the screwdriver 12\% further than \namenospace-Sampled Fixed Sequence and 35\% further than CSVTO-Sampled Fixed Sequence.
This is because it is not always necessary to regrasp all fingers before executing another turn, as is done in the data.
In addition, it is often possible to execute multiple turn modes in a row, even though this does not occur in the data.
While the prior could be altered, what is significant is that we are able to generate trajectories that outperform the data on which $M$ is trained. 
We find the Diffusion Policy turns the screwdriver slower than \namenospace, limiting how far it turns.

In addition, replanning the contact sequence is beneficial as error in the diffusion, even with variability modeling, can lead the planner to overestimate screwdriver turning and lead to a sub-optimal plan which does not reach the specified $\mathbf{o}_G$.
Without replanning, as there is a fixed goal for the search, it is possible for A* to terminate before expanding to the maximum depth, leading to a shorter overall trajectory.

We show that our variability propagation method leads to 23\% further turning than the no variability propagation ablation with lower variance in the yaw angle.
Without any variability propagation, low-quality samples from $M$ can lead the planner to choose lower-quality contact mode sequences.
We can see the utility of $\Psi$ by comparing the max-likelihood and no variability propagation methods.
Choosing the maximum likelihood sample outperforms choosing a random sample, indicating that $\Psi$ helps select higher-quality trajectories.
However, \name outperforms the maximum likelihood ablation.
We believe this is due to overly trusting the $\Psi$ output, which as a learned model also has error.
In addition, $\mathcal{D}$ can contain trajectories that do not satisfactorily perform the task due to failures of trajectory optimization.
Diffused trajectories could therefore be similar to the dataset but low-quality.
Storing a set of particles allows us to better account for low-quality samples from $M$ and errors in $\Psi$.

\looseness-1
Diffusing initializations for CSVTO takes 3.6 s on average and CSVTO takes 7.2 s per step, while Diffusion policy takes 1.1 s.
Average A* planning time is 46.8 s for \namenospace, 13.7 s for ablation (3), 14 s for ablation (4), and 16.4 s for ablation (5).
Expanding an edge with CSVTO would take approximately 20x long as using $M$, motivating the use of $M$ in the search.

\vspace{-.3cm}
\subsection{Real Screwdriver Turning}
\vspace{-.1cm}
We also perform the screwdriver turning task in the real world (shown in Fig.~\ref{fig:title}), using the same specifications for the A* search and using the same models for $M, \Psi$.
$\mathbf{o}_t$ is estimated using Aruco tags on the screwdriver.

\begin{figure}[t]
    \centering
    \includegraphics[scale=.6]{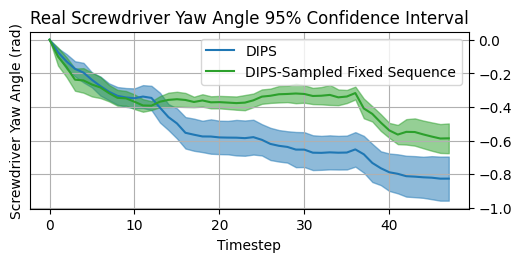}
    
    \vspace{-0.4cm}
    \caption{Real screwdriver manipulation results for valid executions over 10 trials. \name (10/10 valid) outperforms the ablation (8/10 valid) by 41\%.
    }
    \label{fig:real_results}
\end{figure}

For this task, we run \name and \namenospace-Sampled Fixed Sequence for 10 trials, executing 4 modes, to demonstrate the utility of the contact planning over the prior in the real world.
Due to imperfect modeling in our trajectory optimization and limitations of the hardware, we alter the force initializations sampled from $M$.
Directly initializing with the diffused forces leads CSVTO to output forces that are too low to turn the screwdriver, as part of $J_{smooth}$ is a regularization on force magnitude.
We initialize as in \cite{fanrolling} for the thumb and middle fingers in the turn mode.
The diffused trajectories used in A* are not altered.
This sim-to-real gap can be addressed through more advanced modeling of the forces in the trajectory optimization.

As shown in Fig.~\ref{fig:real_results}, \name outperforms the ablation, turning 41\% further.
\name plans 25\% more turning modes than the ablation, reducing unnecessary regrasps.
However, due to perception and execution error and possibly the change in CSVTO initialization, we turn 74\% as far as in simulation.
These errors also lead to the ablation dropping the screwdriver twice while \name does not drop it.
When regrasping the index finger, perception and execution errors can lead the thumb and middle fingers to be ill-positioned to support the screwdriver.
We find \name plans index regrasps only twice across 10 trials as they are less likely to enable further turning than thumb/middle regrasps.
Every ablation trial includes an index regrasp due to the structure of $p(C)$.
We believe these errors in regrasping help explain the gap in performance between \name and the ablation beyond planning more turns.

Diffusing initializations for CSVTO takes 3.1 s on average, while CSVTO takes 16.1 s per step.
We use a higher CSVTO budget on hardware.
Each A* planning call takes 10.1 s on average.
A* planning times are lower than in simulation as we only execute 4 contact modes.
We find the planning time can be higher for later contact modes in simulation.

\vspace{-.2cm}
\section{Conclusion}
\vspace{-.1cm}
We presented \namenospace, a planning method for contact-rich manipulation that combines generative modeling and search methods.
We constructed an approximation to a trajectory optimizer by using a diffusion model trained on optimized trajectories. Then, using a particle-based representation, we reasoned about variability in the diffusion to plan contact sequences that outperformed those in the training set.
\name outperformed ablations and baselines, including on a challenging hardware screwdriver turning task.
As our method requires defining all considered contact modes in the data generation and search, future work could investigate methods to automatically generate task-relevant contact modes.





\bibliographystyle{IEEEtran}
\bibliography{ref.bib}

\end{document}